# Automated Training and Maintenance through Kinect

Saket Warade[1], Jagannath Aghav[1], Claude Petitpierre [2] and Sandeep Udayagiri[3]

[1]Department of Computer Engineering and Information Technology,
College of Engineering, Pune (COEP), India
{waradesp10,jva}.comp@coep.ac.in
[2] Networking Laboratory
Swiss Federal Institute of Technology in Lausanne
CH-1015 Lausanne EPFL, Switzerland
claude.petitpierre@epfl.ch
[3]John Deere Technology Centre India
UdayagiriSandeepC@JohnDeere.com

## Abstract

*In this paper, we have worked on reducing burden on mechanic involving complex automobile maintenance activities that are performed in centralised workshops. We have presented a system prototype that combines Augmented Reality with Kinect. With the use of Kinect, very high quality sensors are available at considerably low costs, thus reducing overall expenditure for system design. The system can be operated either in Speech mode or in Gesture mode. The system can be controlled by various audio commands if user opts for Speech mode. The same controlling can also be done by using a set of Gestures in Gesture mode.*

*Gesture recognition is the task performed by Kinect system. This system, bundled with RGB and Depth camera, processes the skeletal data by keeping track of 20 different body joints. Recognizing Gestures is done by verifying user movements and checking them against predefined condition. Augmented Reality module captures real-time image data streams from high resolution camera. This module then generates 3D model that is superimposed on real time data.*

## 1. Introduction

Training in automobile industry is very important in order to perform maintenance and repair activities. Currently, these training sessions are held in centralized workshops with the guidance of supervisor. The supervisor has to look for every step manually, thereby increasing his workload. Also, he cannot look after every operation that is being carried out by the mechanic. In case of any complex maintenance / repair work which is not a regular one, supervisor have to be present physically to location of operation. This causes delay in operation considering the fact that availability of supervisor cannot be guaranteed to all the locations. Our work is a step towards eliminating these hurdles and automating such activities.

We are tracking the human body joints to keep track of human body. Also, we have used Augmented Reality to track the objects like parts that are used in operations. Earlier, to perform such operations, expensive cameras and sensors were required. But now, with the launch of

Kinect, the same accuracy cameras are available at much cheaper rate. Now, the developers and designers are provided with a small, portable device capable of sensing with high accuracy rate. Using the Depth cameras installed on front panel, this device can get the much required 'z' parameter to calculate the distance between object and the sensor. The Kinect is an excellent addition to the NUI and gestural technologies. The virtual training system is able to interact well with the technician through gesture and speech recognition systems. It is also able to guide him through the complete steps in the training and also able to identify and track his gestures to ensure he is following the right steps and correct him when necessary.

This paper is organized as follows: Section 2 gives brief information about previous work on motion capture, depth sensing and augmented reality. Also few Kinect Identification techniques are also discussed. Section 3 explains how the proposed system takes advantage of Kinect sensors and Augmented Reality to track the movements of mechanic working in workshop. Section 4 gives detailed design of Kinect system including the control modes, database used and an algorithm for 'Rightsweep' gesture with is used in gesture control mode.

## 2. LITERATURE REVIEW

Motion capture and depth sensing are two emerging areas of research in recent years. With the launch of Kinect in 2010, Microsoft opened doors for researchers to develop, test and optimize the algorithms for these two areas. Leyvand T [2] discussed about the Kinect technology. His work throws light on how the Identity of a person is tracked by the Kinect for XBox 360 sensor. Also a bit of information about how the changes are happening in the technology over the time is presented. With the launch of Kinect, Leyvand T expects a sea change in the identification and tracking techniques. The authors discussed the possible challenges over the next few years in the domain of gaming and Kinect sensor identification and tracking. Kinect identification is done by two ways: Biometric sign-in and session tracking. They considered the face that players do not change their cloths or rearrange their hairstyle but they do change their facial expressions, gives different poses etc. He considers the biggest challenge in success of Kinect is the accuracy factor, both in terms of measuring and regressing. Key prospect of the method is they are considering a single depth image and are using an object recognition approach. From a single input depth image, they inferred a per pixel body part distribution.

Jamie Shotton [1] took the advantage of the depth images for human pose recognition. The pixels in depth images indicate the depth in the image data and not any intensity or color information. This helps in calculating the 'z' (depth) parameter. They labelled the body parts according to body part position with respect to the camera. The body is recognized as a set of 31 different labelled parts. They have recognized the body in the set of 31 different labelled parts. Machine Learning is performed by using classification techniques. With decision trees and forests, training is provided to machine.

Christan Plagemann and Varun Ganpathi [11] proposed a feasible solution for identification and localization of human body parts while dealing with depth images. The greatest advantage of this method is that the output can be used directly to infer the human gestures. It can also be used to study and test other different algorithms which involve the detection of human body parts and depth images. The system identified and localizes the body parts into 3D space. To obtain the results the machine is provided training data and classification technique is used to differentiate between two body parts and also between body part and other similar objects. The test results show that the system is able to identify body parts in different conditions and in different locations.

Depth imaging refers to calculating depth of every pixel along with RGB image data. The Kinect sensor provides real-time depth data in isochronous mode[18]. Thus in order to track the movement correctly, every depth stream must be processed. Depth camera provides a lot of advantages over traditional camera. It can work in low light and is color invariant [1] the depth sensing can be performed either via time-of-flight laser sensing or structured light patterns combined with stereo sensing [9]. The proposed system uses the stereo sensing technique provided by PrimeSense [21]. Kinect depth sensing works in real-time with greater accuracy than any other currently available depth sensing camera. The Kinect depth sensing camera uses laser beam to predict the distance between object and sensor. The technology behind This system is that the CMOS image sensor is directly connected to Socket-on-chip [21]. Also, a sophisticated deciphering algorithm (not released by PrimeSense) is used to decipher the input depth data. The limitations for depth cameras are discussed by Henry [9].

J. Gall [7] proposed a method to capture performance of a human or animal from multi-view video sequence. Their method works even for a very rapid and small movement done by the object (animal or human). Once provided with the multi-view image sequences, they tracked the skeletal as well as the surface area of the body. Optimization in skeletal pose is done in order to find the body poses in current frame. The approach requires an articulated template model to superimpose the skeletal on the body of animal. Local optimization reduces the computational burden for global optimization since it can be performed in lower dimensional subspaces to correct the errors. Body of the person is identified by tracking bone joints. Implementation is done with the help of Skeleton based Pose estimation and Surface estimation. The Skeleton based pose estimation used 3D constraints whereas Surface estimation uses 2D constraints. They tested the method on large number of objects with different apparels and found successful.

A motion capture system is a sensors-and-computers system that recovers and produces three-dimensional (3-D) models [7] of a person in motion. It is used in military [3], entertainment, sports etc. for validation purpose. In motion capture sessions, movements of one or more actors are sampled many times per second, although with most techniques motion capture records only the movements of the actor, not his/her visual appearance. To capture motion of a person, the very first step is to check identify the person. J Gall [7] did this by first obtaining the skeletal view of body. Skeletal view is obtained via capturing the body by synchronized and calibrated cameras. Local optimization is performed in order to reduce the complexity for Global optimization. The skeletal view is then mapped with the deformed surface available in database. 3D model is then formed by performing Global optimization on deformed surface. The motion capture algorithm given by J Gall [7] is very efficient for capturing and processing motion of an object. It has observed that the segment parameters of human body are indispensable to compute motion dynamic which causes inaccuracies. Kinect device used in our system is powered by both hardware and software. It does two things: generate a three-dimensional (moving) image of the objects in its field of view and recognize humans among those objects [16].

Robert Y. Wang [12] successfully tracked the hand wearing a color glove with pre-defined custom pattern. To solve the issue of expensive and difficult to port available tracking systems, he proposed a system which is easy to use and is inexpensive. Various ways were explored to track a hand including marker based motion tracking and hand tracking with color markers. They performed the pose estimation with the help of customized glove design and sampling databases. The glove design contains pasting dense patterns of coloured markers for tracking. Colour glove is used for hand tracking because the bare hand gives same position on palm up and palm down whereas if we use a color glove, then we have different positions of the pattern for different poses and hence it improves the tracking accuracy. The training dataset contains more than 18,000 finger positions, hence making it possible to track the hand more accurately to the fingers level.

Henderson and Feiner [3] explored Augmented Reality system design for Maintenance and Repair work. They provided a state-of-the-art prototype that supports military mechanics conducting routine maintenance tasks inside an armoured vehicle, turret. They provided the interaction between system and the mechanic via Augmented Reality concept. The mechanic wears a special type a display glasses which are used to display the instructions. The system is controlled by a wrist-worn hand-held device running on Android. An android application is written using open source Android APIs and Android SDK released by Google early in 2009. The application provides five forms of augmented reality content to assist mechanic. The content includes 3D and 2D arrows, Text instructions, labels, a close up view and 3D model of tools (e.g. a screwdriver). The arrows are used in such a fashion that it becomes denser when the mechanic is moving towards required tool and becomes fader if he is moving away from it. A small animation plays when mechanic reaches to the tool and arrow disappears.

Jun-Da Huang [35] used gesture tracking capability of Kinect in physical rehabilitation system viz. Kinerehab. In this system, gestures are used to find out whether the rehabilitation has reached a particular standard and whether the movements of students are correct or not. An interactive interface using Kinect also enhances student's motivation, interest and perseverance with rehabilitation.

Hand gesture detection is an important aspect of HCI. The authors of [31] used Kinect for hand detection and gesture recognition. But typical resolution of 640*480 for Kinect sensor provides problem in recognition of hand. It was eliminated using a novel shape distance metric called Finger-Earth Mover's Distance to measure the dissimilarities between different hand shapes[32].

## 4. PROPOSED SYSTEM DESIGN

Our system implements Augmented Reality using processing capabilities of Kinect. The system consists of 4 major components as Tracking Device, Processing Device, Input Device and Display Device. We use Kinect as a Tracking device as shown in figure 1. It contains three sensors for processing of depth images, RGB images and voice. Depth camera and Multi-Array Mic of Kinect are used to capture Real-Time image stream and audio data respectively. Depth sensor is used to obtain the distance between sensor and tracking object. The input device to our set-up is a high definition camera which is used to get input image stream and run as the background to all Augmented Reality components. On this background stream, we superimpose event-specific 3D models to provide virtual reality experience. The processing Device, consisting of Data Processing Unit, Audio Unit and software associated with it takes care of which model to superimpose at which time. Processing Unit passes the input video stream and the 3D model to display device for visualization purpose.

The proposed system tracks the movements of mechanic by processing the skeleton data. First, the body joints are identified and later, bones are drawn by joining appropriate joints. Using Kinect we identify 20 joints of human body and track their positions. We detect a motion by considering the difference between two consecutive frames. We identify the motion in a particular direction by taking difference between particular direction parameter(x, y or z).

While performing maintenance work, if the mechanic is going out of range of the specified area, the system sounds an alarm. This enables the supervisor to check whether mechanic is moving out without completing the allocated work.

**Figure 1.  System Architecture**

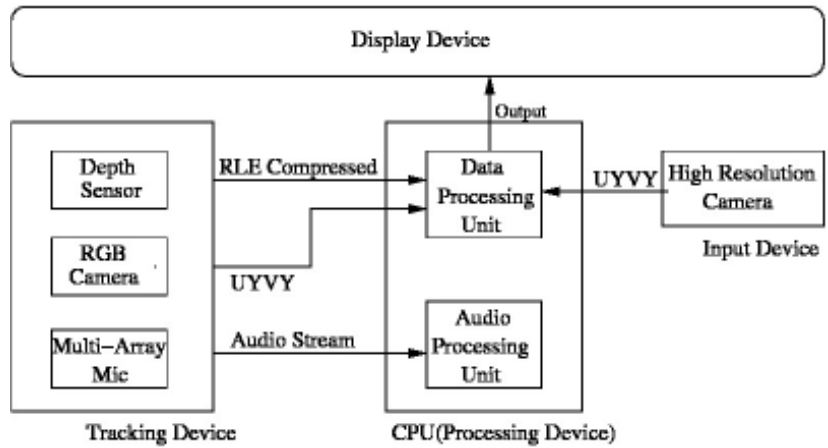

The system is controlled with the help of audio commands as well as gesture inputs. Table 1 gives list of audio commands and their details. The system flow is given in System Flow:

```
-------------------------------------------------------------------------------------
 System Flow
-------------------------------------------------------------------------------------

 1: Start
 2: Identify Position of operation
 3: Locate mechanic in specified area
 4: Guide mechanic to reach to position of operation by audio commands
 5: Make sure that mechanic is ready
 6: Display (Visualize) next Instruction
 7: Wait for signal from mechanic
 8: Require More Details?
 9:      Run pre-captured animation
10:      goto Step 6
11: Repeat Instructions?
12:      Repeat Instruction
13:      goto Step 6
14: Done?
15:      goto Step 5
16: Repeat 6-9 until all activities are performed
17: Verify the result
18: Stop
-------------------------------------------------------------------------------------
```

The system checks whether the mechanic is performing with correct tool or not. Since co-ordinates of every tool is fixed, We can obtain the difference of two depth images, one taken before start of operation and the other one while performing. The difference data is then compared with the shape of the tool the mechanic is supposed to use. If both the shapes are found to be equal, we conclude that correct tool is selected. If correct tool is selected, the mechanic is notified by green signal by the system. If wrong tool is selected, he is notified by Red signal. When the mechanic signals the system, he will visualize the next step on the display provided. The system understands the signals with the help of pre-defined gestures and audio commands. When the mechanic says "NEXT COMMAND" and waves his right hand from rightmost position to leftmost position, the Kinect system understands that the user wants to move on to next instruction. All the visual effects are processed by AR system. Once all plug-ins are loaded, AR system adds particular event specific model to the screen-graph and provides Virtual Reality experience. After that, according to signal received, the event specific model is loaded and unloaded.

There are pre captured sessions involving the experts for every maintenance and repair activity. Animations or light-weighted (compressed) videos are prepared according to experts actions performed in centralized workshops. These Animations/Videos are played if mechanic chooses for "More Details" option. The system also "Repeat" the instructions if mechanic wants to visualize the step information again. Also the mechanic can visualize "Previous Instructions" if he wants to cross check the work done. The system keeps track of this movement and marks that activity as "Completed" or "Current" or "Yet to start". Table 1 gives the list all audio commands used to implement the system.

Motion or movement is detected by considering difference between two frames. The Kinect system is very efficient in tracking skeletal of human body. The tracking is done by identifying different body parts and Joints. For the tracking purpose, the Kinect system considers the fact that human body is capable of giving enormous range of poses.

## 5. DESIGN OF KINECT SYSTEM

The Kinect system plays an important role in working of overall system. This system works as tracking unit for the Augmented Reality System. This system uses some of most exciting functionalities of Kinect such as skeletal tracking, joint estimation and Speech recognition for a human body. Skeletal tracking is useful for determining the user's position from Kinect, when user is in frame, which will be used for guiding him through assembly procedure. Also, it helps in gesture recognition. This system guides the user through complete assembly of product using speech and gesture recognition. The assembly of product includes bringing together individual constituent parts and assembling them as a product.

There are two assembly modes for this system, Full Assembly and Part Assembly. In Full Assembly mode, Kinect will guide technician on how to assemble a whole product sequentially. This mode will be useful when whole product has to be assembled. In Part Assembly mode, technician has to select a part to be assembled and then Kinect will guide him on how to assemble a selected part. When assembly of that part is completed, technician can select another part or quit. This mode will be useful when a part/parts needs to be assembled.

The system has been developed to work in 2 modes, Speech Mode and Gesture mode. The choice to select a mode has been given to user based on his familiarity to system and convenience to use it. If user has opted for speech mode, he has to use voice commands to interact with the system and system will guide him through voice commands. On the other hand, if user has opted for gesture mode, he has to use gesture to interact with the system and system will guide him through voice commands. The 'START' command is used in both modes to initiate the system. After system initiation, user will select a speech mode or gesture mode and will continue working in the same.

### 5.1. System Operation Modes

As mentioned in section 4, the system can be operated in either speech mode or in Gesture mode. In Speech mode, the system can be controlled totally by using audio commands whereas in Gesture mode, various gestures are developed to control the system.

### 5.1.1. Speech Mode

In this mode, various audio commands are used to control the operation of system. Table 1 enlists all the audio commands used to control the system behaviour. 'START' is used for system initialization. When this command is recognized, mode selection screen will display as shown in figure 2, asking user to select mode of his interest. After 'STOP' command, the system will stop working and will show an initial screen. If 'START' command is recognized after this, then system will initialize again.

'NEXT INSTRUCTION' is used only in full assembly mode. When assembly of one part is completed and want to initialize for the next part, this command is given. If assembly of any part is not completed and this command is given, then it will start assembling next part. Similarly, 'PREVIOUS INSTRUCTION' is the command used to go to previous step in the assembly. Clearly, this command works only in Full Assembly mode of operation.

**Table1.** List of Audio Commands

| Sr. | Command | Description |
|---|---|---|
| 1 | Start | Starts the system working |
| 2 | Pause | Pause the system working |
| 3 | Next Instruction | Visualize next instruction on display |
| 4 | More Details | Run pre-captured animations or videos |
| 5 | Repeat Instruction | Re-Visualize the current instruction on display |
| 6 | Previous Instruction | Visualize previous instruction on display |
| 7 | Resume | Resume system working from Pause |
| 8 | Stop | Stops the system working |

'PAUSE' is used to pause the system working. When system is paused, no operation takes place until 'RESUME' command is recognized. At the end, 'STOP' command stops the system and from this point, System can be 're-started'.

**Figure 2.** Mode Selection Screen

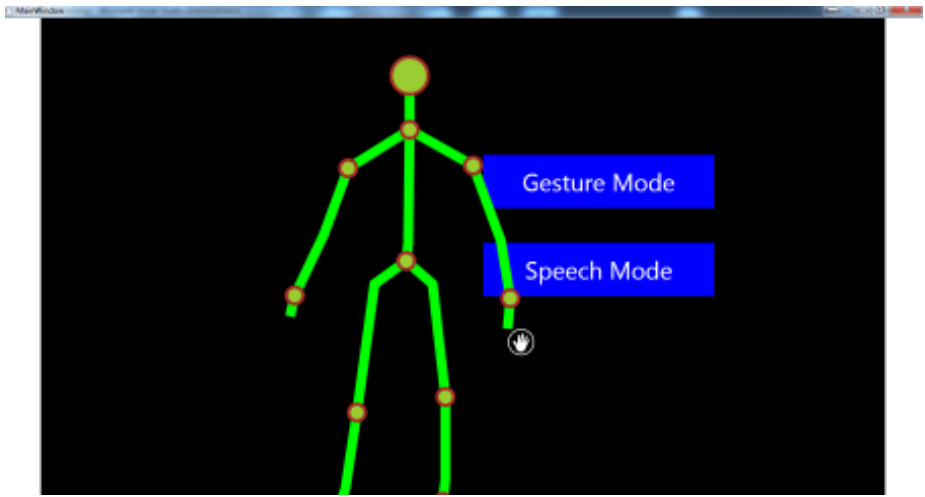

### 5.1.1. Gesture Mode

In this mode all the operations except start are controlled by various gestures. Table 2 gives the full list of gestures developed and used for controlling this system. The system operation starts only by 'START' audio command. From there on, gestures are used everywhere to perform all the operations.

'RIGHTSWEEP' is used only in full assembly mode. When assembly of one part is completed and want to initialize for the next part, this gesture needs to be performed.

**Table 2.** List of Gestures

| Sr. | Gesture | Corresponding Audio Command | Operating Mode |
|---|---|---|---|
| 1 | Hands Up | Pause | FULL, PART |
| 2 | Right Sweep | Next Instruction | FULL |
| 3 | Zoom-in | More Details | FULL, PART |
| 4 | Zoom-out | Repeat Instruction | FULL, PART |
| 5 | Left Sweep | Previous Instruction | FULL |
| 6 | Hands Forward | Resume | FULL, PART |
| 7 | Hands Up | Stop | FULL, PART |

If assembly of the part not completed and this gesture is recognized, then also it will start assembling next part. Figure 3 shows how 'RIGHTSWEEP' is used to obtain next instruction. Left side upper corner gives skeletal and RGB view of user. In the left, we have image of the part we have to fix (Details of operation to be performed).

**Figure 3.** RightSweep Gesture

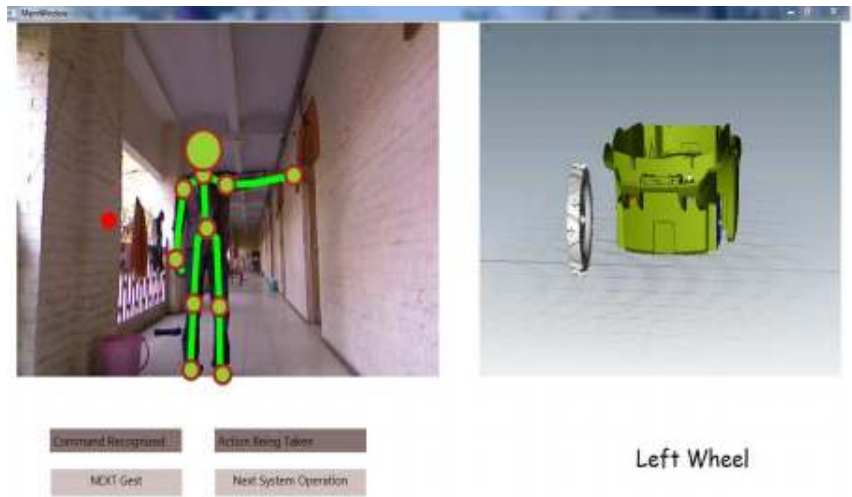

## 5.2. System Workflow

This section gives the overall system workflow for Kinect System. Figure 4 shows the workflow diagram for Kinect system defined and explained in section 4. The workflow is gives all possible states of the system. Green lines indicate the execution of either 'NEXT' audio command or 'RIGHT SWEEP' gesture. Similarly, Purple lines indicate execution 'BACK' audio command or 'PREVIOUS INSTRUCTION' gesture. Whereas, Red lines indicate execution of 'STOP' audio command or 'HANDS UP AND FOLDED' gesture.

Once the system is started, it prompts user to select either gesture mode or speech mode. After this all the input commands will vary according to selection of the controlling mode. After this, assembly selection window appears. If user selects PART assembly, then again system asks for part selection.

**Figure 4.** System Workflow

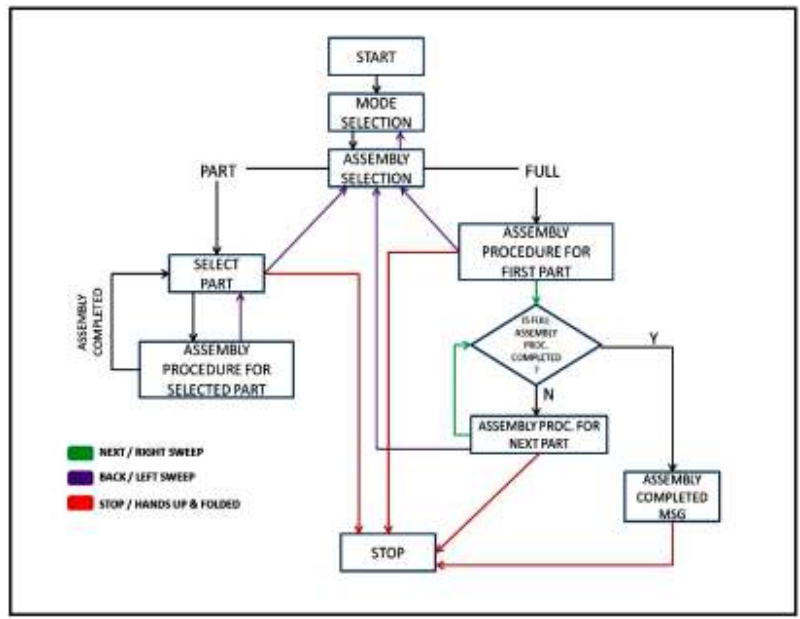

In FULL assembly, the systems performs all operations sequentially one after the other. If user wants to switch to next operation, he is provided this facility with the help of NEXT INSTRUCTION audio command or RIGHTSWEEP gesture. During all these operations, if STOP command is recognized in speech mode or HANDS UP AND FOLDED gesture is recognized in gesture mode, system will stop its functioning. Once all the operations are performed, the system automatically executes STOP and exits.

## 5.3. XML Database

The system uses xml file as database input. This xml file contains all the data related to system operation like operation location and maintenance task ids. Figure 5 shows a sample database for lift and Put operation performed with this developed and implemented Kinect system. The database contains all the information about the part which is to be used for operation. The information is extracted from the *XML* file once user clicks the button associated with part selection.

**Figure 5.** XML database entry

```
-<Part>
    <id>1</id>
    <Part_name>Right Wheel</Part_name>
    <Lift>
            <X>-1.0</X>
            <Z>3.7</Z> </Lift>
    <Put>
            <X>0.4</X>
            <Z>2.0</Z> </Put>
    <Image1>RightWheelLift.jpg</Image1>
    <Image2>RightWheelPut.jpg</Image2>
    <Commands_Lift>Lift Wheel</Commands_Lift>
    <Commands_Put>Fix Wheel</Commands_Put>
    <videoPath>right_wheel_video.avi</videoPath>
</Part>
```

### 5.4. Gesture Recognition

Gesture Recognition usually consists of processing of various images. However, here, we have done it quite smartly by using joint tracking capability of Kinect. We have tracked the joints of user to recognize various gestures. Some of the gestures we have used are implemented simply just by checking the positions of joints (Hands Up, Hands Forward). Algorithm 1 successfully detects 'RIGHTSWEEP' gesture. This gesture we have used for 'NEXT INSTRUCTION'.

In this algorithm, we take input as *gestureperiod* and *frames per second (fps)* value. The *gestureperiod* indicates the time for which you have to check for you gesture and *frames per seconds* indicate how many frames are expected to be processed for each second. We have tested this algorithm for *gestureperiod*. All the data is stored in *localhistory*<> list. *Start* is the starting position of the joint we are tracking (HandRight in this case). This position is calculated as per the *gestureperiod* and *fps*. *index* represents the value of data we should perform all the operation.

The algorithm takes care that if any position is not suitable to perform any action, it directly returns *false* thus declaring that this particular gesture is not possible to happen. Whenever if discard all rejection conditions, it checks for favourable conditions one by one and rejects if any of the condition fails. Finally whenever the right hand covers the distance equal to *reference*, we conclude that the gesture is recognized.

**Algorithm 1.** RightSweep Gesture Recognition

```
Algorithm 1 Right Sweep Gesture Recognition
 1: procedure RIGHTSWEEP(gesturePeriod, fps, localhistory <>)
      /* This procedure checks whether the user sweeps his right hand */
      /* Hand should be in a position between Head and Sholdercenter */
      /* It should be away from body */
 2:   index = gesturePeriod * fps
 3:   start = [localhistory.count − index].Handright.X
      /*index is the variable used to calculate the number of entries */
      /* to check for recognizing Gesture */
 4:   if localhistory.count ≤ index then
 5:     Return false
 6:   end if
      /* If there are no sufficient entries in localhistory to check, */
      /* we can not conclude on Gesture */
 7:   Reference = SholderCenter.Y − Spine.Y
 8:   for all data in [localhistory.count-index] do
      /* Hand is above Head */
 9:     if Handright.Y > Head.Y then
10:       Return false
      /* If Hand is in downward position */
11:     else if Handright.X < Wristright.X then
12:       Return false
      /* Hand is above SholderCenter */
13:     else if Elbowright.X > Handright.X then
14:       Return false
15:     end if
16:     if Handright.X > Sholdercenter.X then
17:       if Elbowright.Y > Handright.Y then
18:         if Handright.Y > Sholdercenter.Y then
19:           if ( then Handright.X > start + Reference)
20:             Return true
21:           end if
22:         end if
23:       end if
24:     end if
25:   end for
26: end procedure
```

## 6. CONCLUSIONS

In this paper, we have discussed how the Kinect sensor is used for Detection and Tracking. We are using Kinect as a tracking device as well as input device for Augmented Reality System. Our work is a step towards automation of maintenance and repair activities for Tractors and other vehicles. The proposed system helps reduce the burden on experts to look into few regular activities. Instead, they can use our system for such activities. This work also simplifies documentation process. The supervisor can keep track of current status of activity from his desk. Also, stepwise verification is possible as the system keeps track of each step. Through the introduction of our system, we will bring new opportunities for mechanical engineering based companies to use Augmented Reality for simplification of their complex tasks. This will add new dimensions to the conventional way of maintenance and Repair activities.

**Authors:**

**Saket Warade** is postgraduate student at the Department of Computer Engineering and Information Technology at College of Engineering, Pune (COEP), India contact him at : waradesp10.comp@coep.ac.in

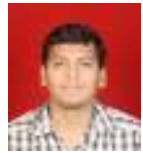

**Dr. Jagannath Aghav** is Professor in the Department of Computer Engineering and Information Technology at College of Engineering, Pune (COEP), India Contact him at : jva.comp@coep.ac.in

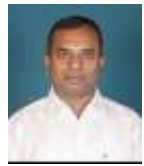

**Claude Petitpierre** is Director at Networking Laboratory, Swiss Federal Institute of Technology in Lausanne CH-1015 Lausanne EPFL, Switzerland. Contact him at: claude.petitpierre@epfl.ch

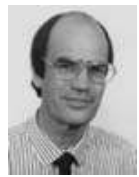

**Sandeep Udayagiri** is Research Analyst in IT Architecture and Innovation Team at John Deere Technology Centre India (JDTCI) Contact him at: UdayagiriSandeepC@johndeere.com

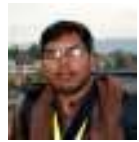